# Nighttime Pedestrian Detection Based on Fore-Background Contrast Learning

He Yao[a], Yongjun Zhang[a,*], Huachun Jian[a], Li Zhang[a], Ruzhong Cheng[b]

[a]State Key Laboratory of Public Big Data, Institute for Artificial Intelligence, College of Computer Science and Technology, Guizhou University, Guizhou, China
[b]North China Institute of Science and Technology, Beijing, China
gs.hyao21@gzu.edu.cn, zyj6667@126.com

**Abstract**

*The significance of background information is frequently overlooked in contemporary research concerning channel attention mechanisms. This study addresses the issue of suboptimal single-spectral nighttime pedestrian detection performance under low-light conditions by incorporating background information into the channel attention mechanism. Despite numerous studies focusing on the development of efficient channel attention mechanisms, the relevance of background information has been largely disregarded. By adopting a contrast learning approach, we reexamine channel attention with regard to pedestrian objects and background information for nighttime pedestrian detection, resulting in the proposed Fore-Background Contrast Attention (FBCA). FBCA possesses two primary attributes: (1) channel descriptors form remote dependencies with global spatial feature information; (2) the integration of background information enhances the distinction between channels concentrating on low-light pedestrian features and those focusing on background information. Consequently, the acquired channel descriptors exhibit a higher semantic level and spatial accuracy. Experimental outcomes demonstrate that FBCA significantly outperforms existing methods in single-spectral nighttime pedestrian detection, achieving state-of-the-art results on the NightOwls and TJU-DHD-pedestrian datasets. Furthermore, this methodology also yields performance improvements for the multispectral LLVIP dataset. These findings indicate that integrating background information into the channel attention mechanism effectively mitigates detector performance degradation caused by illumination factors in nighttime scenarios.*

## 1. Introduction

According to research conducted by psychologists and neuroscientists [1, 2], background information plays a crucial role in object recognition processes. Pedestrian detection is a longstanding challenge in computer vision, with nighttime pedestrian detection presenting even greater difficulties. Unlike daytime pedestrian detection, nighttime pedestrian detection poses challenges due to insufficient natural lighting and the presence of artificial lighting, leading to scenes characterized by uneven illumination, dimness, extreme darkness, and other factors. These conditions often result in issues such as indistinct edges and details of pedestrians, as well as low contrast between pedestrians and the background. Consequently, accurately extracting and distinguishing pedestrian features and contours becomes challenging, ultimately compromising the detector's performance. Nonetheless, applications of nighttime pedestrian detection include video surveillance, assisted driving, and intelligent robotics. Although multispectral pedestrian detection techniques have achieved satisfactory results in these scenarios, their implementation requires integrating expensive sensors into specific devices. Accordingly, the study of single-spectral nighttime pedestrian detection using only RGB images is vital for research and practice in this field.

In the deep learning era, rapid advancements in object detection have led to numerous notable methods. These methods can be classified as multispectral or single-spectral based on their use of infrared images. Thermal imaging cameras can overcome some limitations of color cameras, which struggle to obtain useful information at nighttime, as they are unaffected by lighting conditions. As a result, multispectral methods employ color and thermal images as inputs. Recently proposed nighttime pedestrian detection methods, such as [3, 4, 5], utilize multimodal data. However, the affordability and ease of data acquisition through visible light cameras have led to the proposal of single-spectral nighttime pedestrian datasets, such as NightOwls [6] and TJU-DHD-pedestrian [7], which have facilitated progress in this area. TFAN [8] employs temporal information from the dataset to tackle the occlusion issue in nighttime pedestrian detection. However, this approach is restricted to datasets containing temporal information, necessitating the processing of numerous consecutive frames and adversely affecting real-time performance. EGCL [9] addresses the appearance diversity challenge in nighttime pedestrian detection by constructing example dictionaries for comparative learning on training samples using feature transformations with datasets. However, the size of these dictionaries significantly influences model performance and real-time capabilities. Notably, none of the aforementioned methods consider the issue of suboptimal nighttime pedestrian detector performance resulting from factors such as pedestrian detail loss and low contrast caused by low-light conditions in nighttime settings. Our study tackles these problems by incorporating an attention mechanism.

Drawing inspiration from deep learning interpretability techniques [10, 11, 12] and contrast learning concepts [9, 13], we introduce a novel attention mechanism called Fore-Background Contrast Attention (FBCA). By utilizing the global spatial information of nighttime pedestrian features, FBCA maps the features into a pedestrian feature vector and a background information vector. It then enriches the semantic information of low-light pedestrians through adaptive disparity contrast learning by calculating the vector differences between the two vectors, thereby expanding the disparity between the attention to low-light pedestrian feature channels and background information channels in nighttime images. We develop a new feature fusion module, the Fore-Background Cross Stage Partial Module (FBCsp module), based on FBCA, which facilitates multi-scale feature enhancement of low-illumination pedestrian features at night while minimizing the influence of background information during fusion. We design a new one-stage single-spectral network, the Fore-Background Contrast Network (FBCNet), whose low-illumination pedestrian features not only exhibit a high degree of attention but also significantly differ from background information attention. This distinction arises from two aspects: (1) FBCA considers background information, compressing it into a vector with the same dimensionality as low-illumination pedestrian features. Our method adjusts both dimensions of low-illumination pedestrian features and background information, expanding the difference in attention degree between them and yielding more spatially accurate learned representations. (2) Unlike conventional channel attention mechanisms, our proposed FBCA obtains nighttime pedestrian vectors and background vectors through activation map, allowing for more accurate semantic information while establishing remote dependencies with global spatial feature information. Addressing the challenging issue of accurately extracting and distinguishing features and contours of pedestrians at night, our proposed FBCNet considers the learned fuzzy features as a fusion of two distinct types of information: pedestrian features, influenced by illumination and difficult to discern, and background information unrelated to pedestrians. Subsequently, a pair of vectors is employed to represent the attention of fuzzy features towards pedestrian features and background information, respectively. Due to each feature channel focusing exclusively on one type of information, the pair of vectors exhibits a negative correlation. We then compute the vector difference between these two vectors and utilize it to dynamically adjust the significance of different channels within the fuzzy features. This adaptive adjustment indirectly enhances the channels that emphasize pedestrian features while reducing the emphasis on background information. This approach effectively addresses the challenge of accurately extracting and discriminating the features of pedestrians at nighttime. Additionally, we adopt the neck paradigm from DAMO-YOLO [14] to design a new neck structure that enhances output feature representation and improves model performance. In summary, the main contributions of this work are as follows:

- We introduce the Fore-Background Contrast Attention (FBCA), which adaptively expands the difference in focus between low-illumination nighttime pedestrian features and background information within the network. This allows the model to pay greater attention to pedestrian features with detail loss and low contrast under low-illumination nighttime conditions.
- We design the FBCsp structure based on FBCA. This structure adaptively adjusts the channel feature weights of low-illumination pedestrian features and background information in the feature map after fusing feature maps containing different semantic information. Consequently, pedestrian features and background information are more easily distinguished under low-illumination nighttime conditions, enabling the model to learn more accurate and semantically rich features.
- We propose the Fore-Background Contrast Network (FBCNet), which essentially utilizes background information to adjust the weights of nighttime low-illumination pedestrian features, thereby enhancing the focus on nighttime pedestrian features while weakening background information. This ultimately improves the detector's performance at night.

The remainder of the paper is organized as follows: Section 2 presents related work; Section 3 describes our one-stage single-spectral pedestrian detection methods; Section 4 illustrates the experimental results; and Section 5 concludes the paper.

## 2. Related Work

In this section, we initially provide an overview of pedestrian detection. Subsequently, we examine the relevant literature on single-spectral nighttime pedestrian detection. Lastly, we present a concise summary of the attention mechanism approach.

### 2.1. Pedestrian detection

Prior to the advent of convolutional neural networks, a prevalent solution involved utilizing sliding windows across all potential positions and sizes, an approach inspired by Viola and Jones [15]. Dalal and Triggs [16] proposed the Histogram of Oriented Gradients (HOG) to better characterize pedestrians, while downstream classifiers such as SVM [16] or AdaBoost [17] employed these features for pedestrian detection. To further address pedestrian occlusion issues, Integrated Channel Features (ICF) [18] were introduced for pedestrian detection.

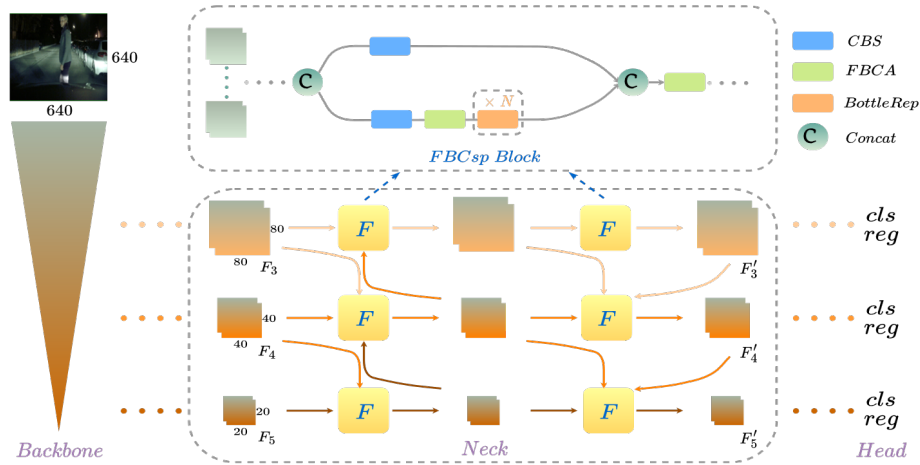

Figure 1. The structure of the proposed Fore-Background Contrast Learning Network (FBCNet) for nighttime pedestrian detection. Based on our proposed FBCA, we designed the FBCsp Block, which can focus on the nighttime pedestrian features and reduce the influence of background information adaptively for the fused features. In order to effectively distinguish and fully exchange the semantic information of nighttime images, we redesign the Neck network, which mainly consists of the FBCsp Block. The image has three outputs after Backbone network: $F_3$, $F_4$ and $F_5$. The three outputs go through the Neck network and have three corresponding outputs, which are $F_3'$, $F_4'$ and $F_5'$. Finally, the three outputs of the Neck network are mapped by the corresponding detection heads to predict the results respectively.

With the rapid advancement of deep learning, particularly convolutional neural networks (CNNs), which facilitated major breakthroughs in image detection, they have become dominant in the research field of general-purpose object detection [19, 20, 21, 22, 23, 24, 25, 26]. This trend has also emerged in the field of pedestrian detection due to the superior feature learning capabilities of CNNs [27, 28, 29, 30, 31, 32, 33]. As Faster R-CNN [34] gained recognition in the object detection field, researchers applied it to pedestrian detection. Subsequently, RPN+BF [35] was proposed to address the issues of low resolution and class imbalance in Faster R-CNN, thereby enhancing pedestrian detection performance. SDS-RCNN [36] then eliminated the imbalance between positive and negative samples by co-learning pedestrian detection and semantic segmentation. Furthermore, owing to the speed advantage and rapid development of one-stage detectors, numerous pedestrian detectors based on one-stage object detection algorithms have been proposed, such as [37, 38, 39], to balance detection accuracy and speed in pedestrian detection.

### 2.2. Single-spectral nighttime pedestrian detection

Since the NightOwls Detection Challenge at the CVPR 2020 Workshop, single-spectral nighttime pedestrian detection has garnered increasing interest from researchers. TFAN [8] employs temporal information between consecutive frames of a dataset to offer additional insight regarding obscured pedestrians, enabling detectors to more accurately detect and localize pedestrians in nighttime scenes. Illumination variation is particularly prominent in night-time scenes, resulting in a greater diversity of pedestrian appearances compared to daytime settings. EGCL [9] enhances the detector's performance in nighttime scenes by transforming the diversity and background information of nighttime pedestrians into a binary classification task, distinguishing pedestrians from backgrounds using a feature conversion module. This module is based on an exemplar dictionary constructed from the dataset. In contrast to EGCL, which addresses nighttime pedestrian detection from the perspective of pedestrian diversity, our approach tackles the challenge of poor model performance due to nighttime illumination. We propose FBCA to enable the FBCNet architecture to concentrate on low-illumination pedestrian features at night from the perspective of illumination variation factors that cause nighttime pedestrians to appear indistinct, low-contrast, and difficult to distinguish.

### 2.3. Attention mechanism

The attention mechanism [40] has proven beneficial for a wide range of computer vision tasks. A notable example is SENet [41], which enhances the network's expressive power by learning the weights of each channel through global average pooling and a multilayer fully connected network, subsequently applying them to the input feature map. CBAM [42] further advanced this concept by introducing spatial information encoding through large-size kernel convolutions. Subsequent research, such as [43, 44], expanded on this idea by employing various spatial attention mechanisms or designing advanced attention blocks.

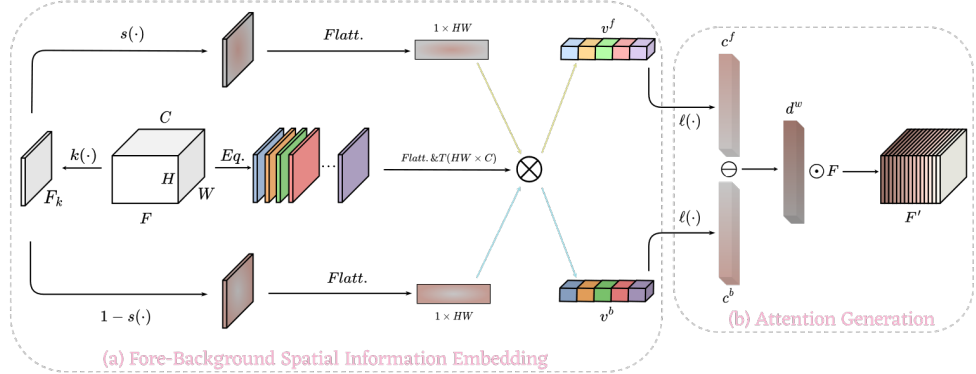

Figure 2. Illustration of FBCA. $k(\cdot)$ is a $k \times k$ convolution with a BN layer. $s(\cdot)$ is a sigmoid function. Flatt.: matrix flattening; $T$: matrix transpose; $\otimes$: matrix multiplication; $\ell(\cdot)$: $MLP\&s(\cdot)$. $\ominus$: vector difference. $\odot$: element-wise product.

Self-attentive networks have gained popularity recently due to their capacity for establishing spatial or channel attention. Classic works, such as [45, 46], both utilize non-local mechanisms to capture different types of spatial information.

In contrast to these attention blocks, our work draws upon the idea of contrast learning [47, 48, 49, 50] to design a Fore-Background contrast attention (FBCA). The FBCA uses global spatial information from the nighttime low-light image feature map to obtain both the nighttime low-light pedestrian feature vector and the background information vector through adaptive transformation. Then, the background information vector guides the nighttime low-light pedestrian feature vector to generate a channel descriptor that enhances nighttime low-light pedestrian attention and weakens background information. This approach ultimately improves the feature representation of the pedestrian detection network in nighttime scenes.

## 3. Proposed method

Our objective is to enhance feature learning for nighttime pedestrian detection, enabling efficient acquisition of low-illumination pedestrian features in nighttime scenes while diminishing the network's focus on background information. This approach addresses the issues of pedestrian detail loss and low contrast caused by low illumination in nighttime scenes. To achieve this, we consider the features learned by the backbone network as a combined problem of nighttime pedestrian feature channels and background information channels. In order to enhance the efficiency and effectiveness of learning nighttime pedestrian foreground features and background information, we introduce a background vector and compute vector differences between the nighttime pedestrian foreground feature vector and the background feature vector. This guides the model towards learning features with rich semantics.

### 3.1. Nighttime pedestrian detection framework

To ensure real-time performance, we select a one-stage detector as our baseline model. Additionally, we find that anchor-free detectors outperform anchor-based detectors for nighttime pedestrian detection. Consequently, we adopt YOLOv6L [51] as the benchmark model. To effectively distinguish and exchange semantic information in nighttime images, we utilize the Efficient-RepGFPN [14] paradigm to design a novel Neck network. Based on this, we construct our Fore-Background contrast learning model, as shown in Figure 1. Our model consists of three parts: the backbone network, the Neck network, and the detection head. The backbone network (e.g., ResNet [52] or VGG [53]) extracts nighttime image features, and for the input of $H \times W$ nighttime pedestrian images, outputs three features $F_3$, $F_4$ and $F_5$ with different semantic information, spatial scales downsampled by 8, 16 and 32, respectively. The Neck network primarily comprises FBCsp modules. Nighttime pedestrian features at different spatial scales must be adjusted to the same spatial dimension through upsampling or downsampling before processing through the FBCsp module of the Neck network. The FBCsp module combines adjusted features containing different semantic information using the concatenate operation and then compresses and fuses channel information through a $1 \times 1$ convolution. Owing to illumination variations in nighttime scenes, the backbone network often learns inadequate semantic information for pedestrians impacted by illumination changes, such as those exhibiting lost detail or low contrast relative to the background. The FBCA embedded in the FBCsp module treats compressed nighttime pedestrian features as a combination of different nighttime pedestrian feature channels and background information channels. FBCA employs the global spatial information of nighttime pedestrian features to map features into nighttime pedestrian feature vectors and background vectors representing channel importance. By computing vector differences between the nighttime pedestrian feature vector and the background vector, the model focuses on contrastive learning between nighttime pedestrian features and background information. This difference contrast learning allows the model to further distinguish the semantic information of nighttime images affected by detail loss, low contrast, and color distortion caused by nighttime illumination changes. The newly designed Neck further enhances semantic information exchange, allowing the model to fully amplify the semantic information within the Neck network. The Neck network generates three outputs: $F'_3$, $F'_4$ and $F'_5$. The detection head employs the Decoupled Head structure from YOLOv6 and ultimately maps the Neck network output into a bounding box.

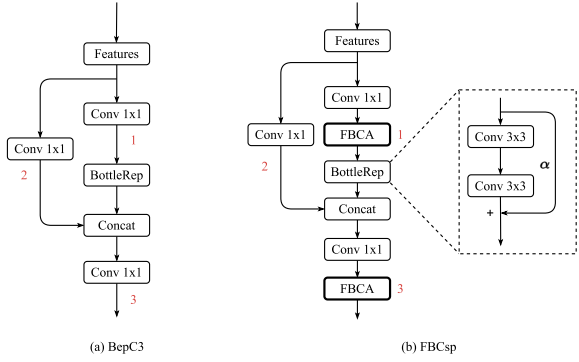

Figure 3. Usage examples. (a) YOLOv6 proposes the BepC3 module; (b) Our proposed FBCsp module.

## 3.2. Fore-Background Contrast Attention Blocks

Our FBCA comprises two steps: Fore-Background spatial information embedding and Fore-Background contrast attention generation. The first step abstracts the pedestrian feature region and background information region under low illumination conditions into two distinct vectors, which represent low-illumination pedestrian features and background information, respectively. The second step calculates vector differences between the pedestrian vector and background vector, which denote low-illumination pedestrian features, to further augment the disparity in network attention to the semantic information and background information of low-illumination pedestrians. Enhancing the distinction between the attention given to low-illumination pedestrian features and nighttime background information allows the network to differentiate low-contrast regions of pedestrians and backgrounds caused by nighttime illumination factors during the learning process. Figure 2 illustrates the proposed schematic diagram for the front background contrast attention mechanism. In the following, we elaborate on the details.

### 3.2.1 Fore-Background Information Embedding

Global pooling is commonly employed for global encoding of spatial information in channel attention (GAP or GMP); however, it compresses the global spatial information into an average or maximum value of spatial information as channel descriptors. Such channel descriptors may not accurately capture the importance of pedestrian feature channels and background channels in low-illumination conditions, where issues such as detail loss, low contrast, color distortion, and shadowing effects arise in nighttime images due to illumination factors. More precise channel descriptors are crucial for networks to concentrate on pedestrian features under low-illumination conditions and minimize attention to background information. To enhance this distinction in attention, we introduce the background information of nighttime images. To enable the attention block to spatially capture long-range dependencies using the precise low-illumination pedestrian features and background information of nighttime images, we employ sigmoid activation maps to decompose the degree of attention given to low-illumination pedestrian features and background information for a specified feature map. Specifically, for part (a) of Figure 2, the feature layer input $F = [f_1, f_2, \ldots, f_C]$ for a given nighttime image has C representing the channel. We initially use a convolution kernel to compress the channel dimension to one dimension and subsequently employ a sigmoid function to map the low-illumination pedestrian region and the background region of the nighttime image. Consequently, the low-illumination equation:

$$F_{map}^f = \sigma(CBLR(F)). \quad (1)$$

Here, $F_{map}^f \in \mathbb{R}^{1 \times H \times W}$, and $CBLR()$ represents the $Conv$, $BN$ and $LeakyReLU$ activation functions, while $\sigma()$ denotes the Sigmoid function. Similarly, the background region of the feature map can be described by the equation:

$$F_{map}^b = 1 - F_{map}^f. \quad (2)$$

Next, the low-illumination pedestrian activation map and background activation map effectively decompose the feature map $F$ into low-illumination pedestrian feature representations and background feature representations, denoted as $v^f$ and $v^b$, respectively. For a given feature map, $v^f$ and $v^b$ can be described as:

$$v^f = F_{map}^f \otimes F^\top, \quad (3)$$
$$v^b = F_{map}^b \otimes F^\top. \quad (4)$$

Here, $F_{map}^f$, $F_{map}^b$ and $F^\top$ are flattened, i.e., $F_{map}^f \in \mathbb{R}^{1 \times HW}$, $F_{map}^b \in \mathbb{R}^{1 \times HW}$ and $F^\top \in \mathbb{R}^{C \times HW}$. $v^f \in \mathbb{R}^{1 \times C}$ and $v^b \in \mathbb{R}^{1 \times C}$. $\otimes$ and $\top$ denote the multiplication and transposition of matrices, respectively.

The aforementioned transformations aggregate features for two spatial regions, specifically nighttime low-illumination pedestrian feature information and background information of the activation map, to obtain a pair of perceptual feature vectors focusing on nighttime low-illumination pedestrians and the background, respectively. These transformations enable our attention blocks to capture long-term dependencies in both the nighttime low-illumination pedestrian region and the background region, effectively compressing the regional information into more accurate one-dimensional vectors. This process enhances the distinction between pedestrian features and background features under low-illumination conditions, assisting the network in accurately localizing illumination-limited nighttime pedestrians.

Table 1. Ablation study for each proposed module on the NightOwls dataset. baseline* indicates that our proposed neck is used. w/o $c^b$ indicates that FBCsp does not consider background information.

| Methods | R | R_S | R_O | All |
|---|---|---|---|---|
| Baseline | 16.28 | 24.30 | 54.17 | 28.89 |
| Baseline* | 15.66 | 22.47 | 53.06 | 27.90 |
| Baseline*+FBCsp (w/o $c^b$) | 14.98 | 24.61 | 55.86 | 27.87 |
| **Baseline*+FBCsp (FBCNet)** | **13.64** | **22.70** | **46.91** | **25.61** |

### 3.2.2 Contrast Attention Generation

As previously discussed, formulas (3) and (4) possess a global receptive field and encode precise low-illumination pedestrian and background information. To capitalize on the resulting expressive representation, we propose a second transformation, referred to as Fore-Background Contrast Attention Generation, corresponding to part (b) of Fig. 2. Our design considers two criteria: (1) it should fully utilize the captured information about low-illumination pedestrian and background areas, accurately highlighting pedestrian regions that are challenging to distinguish due to illumination, and (2) it should effectively capture the relationship between channels to increase the degree of difference between focusing on low-illumination pedestrians and the background. To fulfill these conditions, we employ a simple gating mechanism with sigmoid activation.

$$c^f = \sigma\left(W_2^f \delta(W_1^f v^f)\right), \quad (5)$$
$$c^b = \sigma\left(W_2^b \delta(W_1^b v^b)\right). \quad (6)$$

where $\delta$ denotes the LeakyReLU activation function, $W_1^f, W_1^b \in \mathbb{R}^{C/r \times C}$, $W_2^f, W_2^b \in \mathbb{R}^{C \times C/r}$, $c^f, c^b \in \mathbb{R}^{1 \times C}$, and $r$ are the compression rates. Here, $c^f$ corresponds to the low-illumination pedestrian feature channel, while $c^b$ relates to the background information channel. To further increase the degree of difference between the focus on low-illumination pedestrian features and the focus on background features, we compute the vector difference between $c^f$ and $c^b$:

$$d_w = c^f - c^b. \quad (7)$$

Finally, the adaptively adjusted low-illumination pedestrian

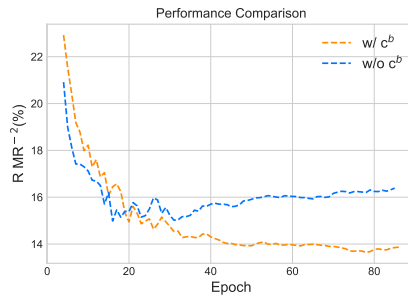

Figure 4. FBCsp module embedded $c^b$ and without embedding $c^b$ performance comparison of the two methods on the NightOwls R subset.

features can be expressed as:
$$F' = F \cdot d_w. \quad (8)$$

Unlike channel attention, our FBCA also accounts for encoding background information. As discussed earlier, the feature representation of the nighttime low-illumination pedestrian region and the background region are concurrently applied to the input features. Each element in the former mapping indicates the presence or absence of low-illumination pedestrians in the corresponding region, while the latter serves to indirectly enhance the attention level toward the low-illumination pedestrian region. This encoding process enables our FBCA to more accurately identify the channels containing low-illumination pedestrian features, thereby improving the overall model's performance in recognizing them. We will provide a detailed demonstration of this in the experimental section.

### 3.3. Implementation

As the aim of this paper is to explore an improved method for enhancing the convolutional features of nighttime pedestrian detection networks, we have designed the FBCsp module based on FBCA as an example to demonstrate the effectiveness of the proposed FBCA block and its advantages over other well-known attention modules. Figure 3 illustrates the integration of Fore-Background Contrast Attention into the CSP structure.

## 4. Experiment

In this section, we initially outline the experimental setup, followed by a series of ablation experiments to showcase the contributions of individual components of the proposed FBCA to performance. Subsequently, we present a comparison of our method with cutting-edge single-spectral nighttime pedestrian detection techniques. Furthermore, to reinforce the effectiveness of our approach, we conduct supplementary experiments using a multispectral dataset. Lastly, we contrast our method with various channel attention-based strategies, highlighting the enhanced performance of FBCA in low-illumination pedestrian detection

Table 2. Comparison of model performance by FBCA for different kernel sizes. Obviously, the appropriate hyperparameters are crucial for good results of our method.

| Methods | K | | R |
|---|---|---|---|
| | 1 | 3 | |
| Baseline | - | - | 16.28 |
| Baseline* | - | - | 15.66 |
| FBCNet | 3 | 3 | 15.69 |
| | 3 | 5 | 15.94 |
| | **5** | **3** | **13.64** |
| | 5 | 5 | 14.41 |

Table 3. On the subset of R $MR^{-2}$ performance and FBCNet's single-image inference speed and batch inference speed on the NightOwls dataset.

| Methods | Speed$^{A100}$ fp16 b1 (fps) | Speed$^{A100}$ fp16 b32 (fps) | R |
|---|---|---|---|
| Baseline | 72 | 380 | 16.28 |
| Baseline* | 57 | 361 | 15.66 |
| **FBCNet** | **42** | **348** | **13.64** |

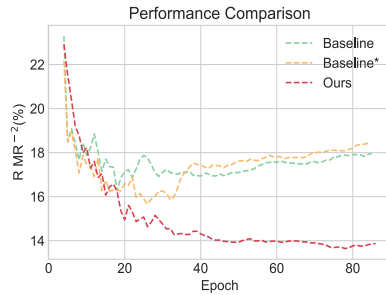

Figure 5. Performance comparison of our proposed method with two baseline models on the NightOwls R subset.

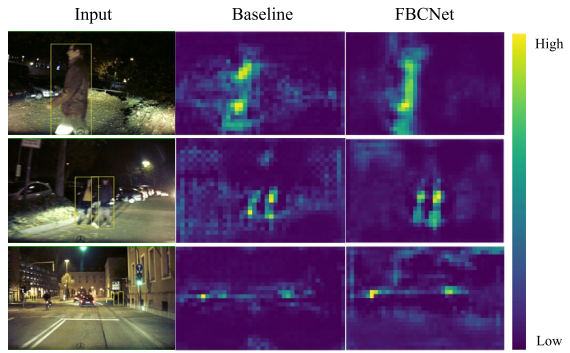

Figure 6. Visualization of the feature maps obtained by both Baseline and FBCNet models using the NightOwls validation set is presented. Both models display feature maps for the entire image, corresponding to ($F'_4$ features).

scenarios.

## 4.1. Experimental setup

### 4.1.1 Datasets

Our proposed method was evaluated on three standard benchmark datasets, including NightOwls [6], and TJU-DHD-pedestrian [7] for single-spectral nighttime pedestrian detection, LLVIP [54] for multispectral nighttime pedestrian detection. NightOwls, a nighttime pedestrian dataset, consists of images captured at night and dawn, with 128k training images, 51k validation images, and 103k testing images. The TJU-DHD-pedestrian dataset, featuring more complex and diverse scenes than NightOwls, presents a challenge for both daytime and nighttime pedestrian detection. This dataset contains 75,246 images, including 373,241 pedestrian labels, and comprises a mix of daytime and nighttime images. It has two distinct subsets: TJU-Ped-campus and TJU-Ped-traffic. In line with Pang et al. [7], we conducted separate experiments on these subsets. Lastly, the LLVIP dataset is a large-scale visible infrared paired low-light visual nighttime pedestrian detection dataset, encompassing various street scenes captured by surveillance cameras. With 15,488 aligned visible infrared image pairs, we used 12,025 pairs for training and 3,463 pairs for testing.

### 4.1.2 Evaluation metrics

According to the standard pedestrian assessment protocol [55], we chose a range of $[10^{-2}, 10^0]$ (denoted as $MR^{-2}$) of the log-average Miss Rate over False Positive Per Image (FPPI) as an evaluation metric. Lower $MR^{-2}$ values indicate better pedestrian detection performance. On the NightOwls dataset, following [6, 8, 9], we reported evaluation results for Reasonable subset (**R**), Reasonable small (**R_S**), Reasonable occluded (**R_O**), and **All**. On the TJU-DHD-pedestrian dataset, with [7, 9] settings, we reported evaluation results for the subsets **R**, **HO**, **R+HO**, and **All**. The LLVIP dataset primarily focuses on multispectral pedestrian detection, and state-of-the-art methods mainly work on the fusion of thermal and RGB images, which is not the focus of our paper. Therefore, we used only RGB images or thermal images and compared our method with the baseline detector on the LLVIP dataset to demonstrate the effectiveness of our method on the multispectral dataset.

### 4.1.3 Implementation details

In our experiments, we utilized the Mosaic [56] and Mixup [57] data enhancement methods. Our network is trained on 1 A100-SXM4-40GB GPU.
**NightOwls:** The image size is 640x640 pixels, and the Stochastic Gradient Descent with Weight Decay (SGDW) optimization algorithm is used to train our model. The initial learning rate is 0.0032, the momentum coefficient is 0.843, and the weight decay is 0.00036. The batch size for training is set to 32. The learning rate strategy is OneCycleLR, with a final OneCycleLR learning rate of 0.12. The number of iterations is 300 epochs. The employed backbone (VGG-16) is initialized using the pre-trained ImageNet model.

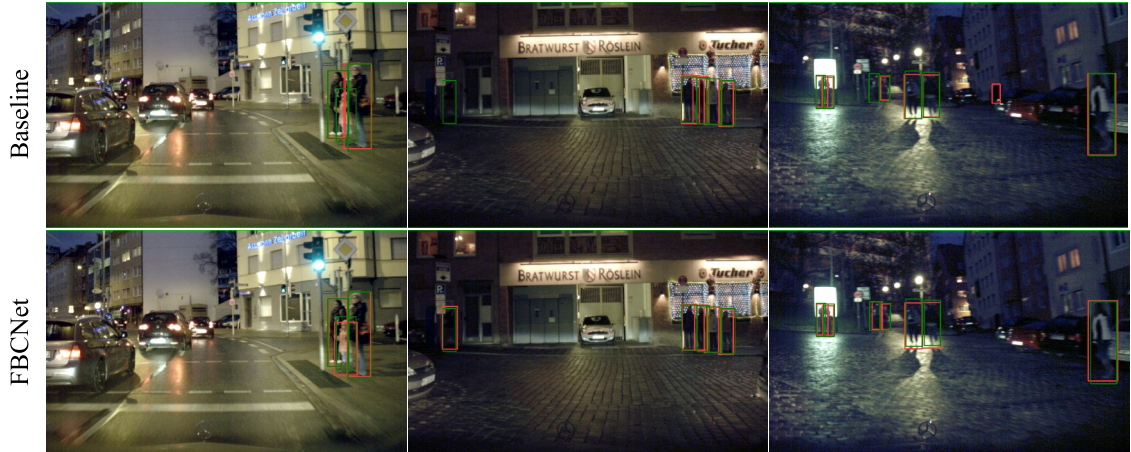

Figure 7. The detection results of FBCNet and Baseline are visualized. Our model successfully detects all nighttime pedestrians, whereas the baseline model struggles to identify small targets, occlusions, and illumination-limited nighttime pedestrians.

Table 4. We evaluate the performance of our proposed FBCNet model with other nighttime pedestrian detection methods on the NightOwls dataset and compare it using the $MR^{-2}$ metric (lower values of this metric indicate better performance).

| Methods | Years | Backbone | R |
|---|---|---|---|
| ACF [58] | TPAMI14 | - | 51.68 |
| Checkerboards [59] | CVPR15 | - | 39.67 |
| Vanilla Faster R-CNN [34] | NIPS15 | VGG-16 | 20.00 |
| RPN+BF [35] | ECCV16 | VGG-16 | 23.26 |
| Adapted Faster R-CNN [60] | CVPR17 | VGG-16 | 18.81 |
| SDS-RCNN[36] | ICCV17 | VGG-16 | 17.80 |
| TFAN [8] | CVPR20 | ResNet-50 | 16.50 |
| EGCL [9] | TIP22 | VGG-16 | 15.93 |
| **FBCNet(ours)** | | VGG-16 | **13.64** |

**TJU-DHD-pedestrian:** The image size is 1280x1280 pixels, and our model is trained using Stochastic Gradient Descent with Weight Decay (SGDW) optimization algorithm. The initial learning rate is 0.01, the momentum coefficient is 0.937, and the weight decay is 0.0005. The batch size for training is set to 8. The learning rate strategy is OneCycleLR, with a final OneCycleLR learning rate of 0.01. The number of iterations is 200 epochs. The model (ResNet-50) is initialized accordingly.

**LLVIP:** The image size is 640x640 pixels, and the AdamW optimization algorithm is used to train our model. The initial learning rate is 0.001, the momentum coefficient is 0.937, and the weight decay is 0.0005. The batch size for training is set to 32. The learning rate strategy is OneCycleLR, with a final OneCycleLR learning rate of 0.1. The number of iterations is 100 epochs. The backbone network (ResNet-101) used is initialized with the ImageNet pre-trained model.

### 4.2. Ablation Studies

**Importance of Background information.** To evaluate the effectiveness of the proposed FBCA in nighttime pedestrian detection, we conducted background vector $c^b$ ablation experiments, and the results are presented in Table 1. We removed the background information from the FBCA to assess the significance of encoded background information. As illustrated in Table 1, we provide a comparison of experiments for our proposed FBCsp module. It is evident that the FBCsp module improves the performance of nighttime pedestrian detection when combined with background information. Figure 4 displays a performance comparison of the two approaches on the NightOwls dataset **R** subset. The experiments indicate that background information embedding is beneficial for nighttime pedestrian detection. baseline* employs the new Neck paradigm we designed.

**Comparison with Baselines (YOLOv6L).** The experimental results presented in Table 1 demonstrate that our FBCNet model significantly outperforms the baseline model (YOLOv6L) across all subsets of the NightOwls dataset. Focusing on the Reasonable (**R**) subset, which is typically employed for performance comparisons, FBCNet enhances pedestrian detection at night, reducing $MR^{-2}$ from 16.28 to 13.64 on the NightOwls dataset. Figure 5 illustrates the considerable performance difference between our proposed method and the baseline model, particularly in the **R** subset.

**Effectiveness of the FBCsp Module.** In general, for the concatenated feature maps, the typical operation involves fusing the features using a 1 × 1 convolution, which serves to compress the channel dimension and reduce subsequent computation. However, this approach neglects the importance of inter-channel relationships. In contrast, the FBCsp module can emphasize low-illumination pedestrian features at night and minimize the influence of background information in nighttime scenes. That is, it indirectly amplifies the difference in the model's attention toward pedestrian features and background features at night, addressing the challenge of distinguishing pedestrians from backgrounds due to illumination changes. The substantial performance gap between "Baseline*" and "Baseline*+ FBCsp" as illustrated in Table 3 and Figure 5, demonstrates the effectiveness of our FBCsp module in nighttime pedestrian detection.

Table 5. We evaluate the performance of our proposed FBCNet model with other pedestrian detection methods on the TJU-Ped-traffic dataset and compare it using the $MR^{-2}$ metric (lower values of this metric indicate better performance).

| Methods | Years | Backbone | Input size | R | HO | R+HO | ALL |
|---|---|---|---|---|---|---|---|
| RetinaNet [24] | CVPR17 | ResNet-50 | 2048×1024 | 23.89 | 61.60 | 28.45 | 41.40 |
| FCOS [61] | CVPR19 | ResNet-50 | 2048×1024 | 24.35 | 63.73 | 28.86 | 40.02 |
| FPN [25] | CVPR17 | ResNet-50 | 2048×1024 | 22.30 | 60.30 | 26.71 | 37.78 |
| CrowdDet [62] | CVPR20 | ResNet-50 | 1400×800 | 20.82 | 61.22 | 25.28 | 36.94 |
| EGCL [9] | TIP22 | ResNet-50 | 1400×800 | 19.73 | 60.05 | 24.19 | 35.76 |
| Baseline* | | ResNet-50 | 1280×1280 | 20.17 | 59.71 | 24.41 | 36.91 |
| **FBCNet(ours)** | | ResNet-50 | 1280×1280 | **19.38** | **57.05** | **23.72** | **35.53** |

**Tuning of hyper-parameters(k).** To demonstrate the influence of hyperparameters on our method's performance, we present the results of nighttime pedestrian detection using various hyperparameter configurations. As the FBCsp module employs the FBCA at positions 1 and 3, we adjusted the FBCA $k(\cdot)$ parameters at these locations using different settings. Table 2 displays the performance of nighttime pedestrian detection in the NightOwls **R** subset under our diverse hyperparameter settings. The experiments indicate that the size of the convolution kernel, $k(\cdot)$, has a substantial effect on model performance.

**Computational complexity.** In order to assess the computational complexity of our model, we evaluated its inference time, as presented in Table 3, which includes both single-image and batch inference speeds. A comparison between the inference times of Baseline and Baseline* reveals that the proposed FBCA introduces negligible additional time, while providing substantial performance enhancements.

### 4.3. Qualitative Study

**Visualization of learned feature maps.** To further assess the efficacy of FBCNet in nighttime pedestrian detection, we conducted a qualitative analysis between FBCNet and Baseline by visualizing their respective feature maps. Figure 6 displays the feature maps for the entire image ($F'_4$ features). FBCNet exhibits enhanced accuracy in detecting nighttime pedestrians, as it concentrates more on nighttime low-illumination pedestrian features while diminishing the focus on background information. This approach expands the attention given to both nighttime low-illumination pedestrian features and background information within the feature space.

**Visualization of pedestrian detection.** In Figure 7, we present a qualitative comparison of nighttime pedestrian detection results obtained using FBCNet and Baseline. This figure illustrates the detection outcomes of both models on three examples from the NightOwls validation set. The baseline model fails to detect nighttime pedestrians who are either very small in size (as seen in the third example) or partially occluded (as in the first example), as well as those situated in areas with limited illumination (as in the second example). Furthermore, the Baseline model generates false detections in regions with poor lighting (as in the third example). In contrast, FBCNet accurately detects all nighttime pedestrians.

### 4.4. Comparison with State-of-the-Art

**NightOwls Dataset.** Our research aims to investigate a novel approach for nighttime pedestrian detection, the FBCNet, and assess its performance by comparing it to state-of-the-art nighttime pedestrian detection methods. We present a comparative evaluation, focusing on the **R** subset, as its results are publicly available. Our experimental findings, detailed in Table 4, indicate that our method outperforms competing approaches by a 2.29-point margin in the **R** subset. This highlights the effectiveness of our proposed FBCA for nighttime pedestrian detection. Furthermore, our method demonstrates exceptional performance in other subsets, further establishing its superiority.

**TJU-DHD-pedestrian Dataset.** The TJU-DHD-pedestrian Dataset comprises two distinct datasets for different scenarios: the TJU-Ped-Traffic Dataset and the TJU-Ped-Campus Dataset. Both datasets include daytime and nighttime images, which pose greater challenges for pedestrian detection, particularly during nighttime. We applied state-of-the-art methods to these datasets and compared the results with our proposed method. As demonstrated by the experimental results in Tables 5 and 6, our method outperforms all other methods on all subsets of both the "Campus" and "Traffic" categories, thereby validating its effectiveness in nighttime pedestrian detection. It is important to note that our method's performance on these two datasets is lower compared to that on the NightOwls dataset, likely due to the cross-domain (day and night) problem.

Table 6. We evaluate the performance of our proposed FBCNet model with other pedestrian detection methods on the TJU-Ped-campus dataset and compare it using the $MR^{-2}$ metric (lower values of this metric indicate better performance).

| Methods | Years | Backbone | Input size | R | HO | R+HO | ALL |
|---|---|---|---|---|---|---|---|
| RetinaNet [24] | CVPR17 | ResNet-50 | 2048×1024 | 34.73 | 71.31 | 42.26 | 44.34 |
| FCOS [61] | CVPR19 | ResNet-50 | 2048×1024 | 31.89 | 69.04 | 39.38 | 41.62 |
| FPN [25] | CVPR17 | ResNet-50 | 2048×1024 | 27.92 | 67.52 | 35.67 | 38.08 |
| CrowdDet [62] | CVPR20 | ResNet-50 | 1400×800 | 25.73 | 66.38 | 33.65 | 35.90 |
| EGCL [9] | TIP22 | ResNet-50 | 1400×800 | 24.84 | 65.27 | 32.39 | 34.87 |
| Baseline* | | ResNet-50 | 1280×1280 | 24.53 | 65.61 | 32.66 | 35.02 |
| **FBCNet(ours)** | | ResNet-50 | 1280×1280 | **23.76** | **64.05** | **31.84** | **34.20** |

Table 8. Comparison of different channel attention methods using the $MR^{-2}$ metric (lower values of this metric indicate better performance).

| Methods | Params(M) | FLOPs(G) | R |
|---|---|---|---|
| Baseline | 34.61 | 291.59 | 16.28 |
| Baseline* | 40.46 | 303.98 | 15.66 |
| + SE | 40.48 | 304.00 | 15.52 |
| + ECA | 40.46 | 304.00 | 15.03 |
| + Coord. Attention | 40.52 | 304.11 | 14.91 |
| **+ FBC Attention(ours)** | **40.50** | **304.10** | **13.64** |

**LLVIP Dataset.** The LLVIP Dataset, a newly constructed large-scale multispectral pedestrian detection dataset, serves to evaluate the effectiveness of our proposed method on multispectral pedestrian detection. A prevalent approach for multispectral pedestrian detection involves fusing thermal and RGB images; however, this is beyond the scope of this paper. Consequently, we exclusively utilized RGB or thermal images from the LLVIP dataset and compared our method with the baseline detector. Table 7 presents the experimental results, indicating performance improvements in both the visible image subset and the thermal image subset. This confirms the generalization and effectiveness of our proposed method in multispectral pedestrian detection.

### 4.5. Comparison with Other Attention Methods

Table 7. We evaluated the generalizability of our proposed FBCNet model on the LLVIP dataset and compared it using the $MR^{-2}$ metric (lower values of this metric indicate better performance).

| Methods | visible | infrared |
|---|---|---|
| Yolov3 [54] | 37.70 | 19.73 |
| Yolov5 [54] | 22.59 | 10.66 |
| Baseline* | 20.90 | 9.02 |
| **FBCNet** | **19.78** | **7.98** |

We contrast our FBCA with alternative channel attention methods. Table 8 presents the impact of incorporating various attentional mechanisms, such as SE [41], ECA [63], Coord. Attention [64], and FBC Attention, on the performance of nighttime pedestrian detection algorithms. Our method surpasses the others, with the metric **R** decreasing to 13.64, while maintaining the number of parameters and floating-point operations virtually unchanged. FBC Attention, an attention mechanism based on fore-background comparison, enhances nighttime pedestrian detection performance by heightening attention to disparities between pedestrian features and background information. This mechanism facilitates the detector's ability to discern nighttime pedestrians from the background, thereby improving nighttime pedestrian detection accuracy. In comparison to other channel attention mechanisms, FBC Attention more effectively manages the attention difference between low-illumination pedestrian features and background information, resulting in superior performance.

## 5. Conclusions

In this paper, we present a Fore-Background Contrast Learning Network (FBCNet) model for single-spectral nighttime pedestrian detection. FBCNet utilizes the FBCA to abstract channels that focus on low-illumination pedestrian features and channels representing background information into two vectors, facilitating adaptive contrast learning. This learning mechanism enables greater emphasis on low-illumination pedestrian features at night, while diminishing the network's focus on background information. Consequently, it allows for effective learning of indistinguishable pedestrians affected by illumination factors in nighttime scenes. In numerous nighttime pedestrian detection experiments, the proposed FBCNet model demonstrates its effectiveness in the task of nighttime pedestrian detection.

# References

[1] Wolfe J M. Guided search 2.0 a revised model of visual search[J]. Psychonomic bulletin & review, 1994, 1: 202-238.

[2] Wolfe J M, Horowitz T S. Five factors that guide attention in visual search[J]. Nature Human Behaviour, 2017, 1(3): 0058.

[3] Zhang L, Zhu X, Chen X, et al. Weakly aligned cross-modal learning for multispectral pedestrian detection[C]//Proceedings of the IEEE/CVF international conference on computer vision. 2019: 5127-5137.

[4] Zhou K, Chen L, Cao X. Improving multispectral pedestrian detection by addressing modality imbalance problems[C]//Computer Vision- ECCV 2020: 16th European Conference, Glasgow, UK, August 23-28, 2020, Proceedings, Part XVIII 16. Springer International Publishing. 2020: 787-803

[5] Cao Y, Luo X, Yang J, et al. Locality guided cross-modal feature aggregation and pixel-level fusion for multispectral pedestrian detection[J]. Information Fusion, 2022, 88: 1-11.

[6] Neumann L, Karg M, Zhang S, et al. Nightowls: A pedestrians at night dataset[C]//Computer Vision–ACCV 2018: 14th Asian Conference on Computer Vision, Perth, Australia, December 2–6, 2018, Revised Selected Papers, Part I 14. Springer International Publishing, 2019: 691-705.

[7] Yanwei Pang, Jiale Cao, Yazhao Li, Jin Xie, Hanqing Sun, and Jinfeng Gong. Tju-dhd: A diverse high-resolution dataset for object detection. IEEE Transactions on Image Processing, 2020. Transactions on Image Processing, 2020.

[8] Wu J, Zhou C, Yang M, et al. Temporal-context enhanced detection of heavily occluded pedestrians[C]//Proceedings of the IEEE/CVF Conference on Computer Vision and Pattern Recognition. 2020: 13430-13439.

[9] Lin Z, Pei W, Chen F, et al. Pedestrian detection by exemplar-guided contrastive learning[J]. IEEE transactions on image processing, 2022.

[10] Zeiler M D, Fergus R. Visualizing and understanding convolutional networks[C]//Computer Vision–ECCV 2014: 13th European Conference, Zurich, Switzerland, September 6-12, 2014, Proceedings, Part I 13. Springer International Publishing, 2014: 818-833.

[11] Zhou B, Khosla A, Lapedriza A, et al. Learning deep features for discriminative localization[C]//Proceedings of the IEEE conference on computer vision and pattern recognition. 2016: 2921-2929.

[12] Selvaraju R R, Cogswell M, Das A, et al. Grade-cam: visual explanations from deep networks via gradient-based localization[C]//Proceedings of the IEEE international conference on computer vision. 2017: 618-626.

[13] Xie J, Xiang J, Chen J, et al. Contrastive learning of class-agnostic activation map for weakly supervised object localization and semantic segmentation[J]. arXiv preprint arXiv:2203.13505, 2022.

[14] Xu X, Jiang Y, Chen W, et al. DAMO-YOLO : A Report on Real-Time Object Detection Design[J]. arXiv preprint arXiv:2211.15444, 2022.

[15] Viola P, Jones M J. Robust real-time face detection[J]. International journal of computer vision, 2004, 57: 137-154.

[16] Dalal N, Triggs B. Histograms of oriented gradients for human detection[C]//2005 IEEE computer society conference on computer vision and pattern recognition (CVPR'05). Ieee, 2005, 1: 886-893.

[17] Bourdev L, Brandt J. Robust object detection via soft cascade[C]//2005 IEEE Computer Society Conference on Computer Vision and Pattern Recognition (CVPR'05). IEEE, 2005, 2: 236-243.

[18] Dollár P, Tu Z, Perona P, et al. Integral channel features[J]. 2009.

[19] He K, Gkioxari G, Dollár P, et al. Mask r-cnn[C]//Proceedings of the IEEE international conference on computer vision. 2017: 2961-2969.

[20] Sun S, Pang J, Shi J, et al. Fishnet: A versatile backbone for image, region, and pixel level prediction[J]. Advances in neural information processing systems, 2018, 31.

[21] Liu W, Anguelov D, Erhan D, et al. Ssd: single shot multibox detector[C]//Computer Vision-ECCV 2016: 14th European Conference, Amsterdam, The Netherlands, October 11-14, 2016, Proceedings, Part I 14. Springer International Publishing, 2016: 21-37.

[22] Redmon J, Divvala S, Girshick R, et al. You only look once: Unified, real-time object detection[C]//Proceedings of the IEEE conference on computer vision and pattern recognition. 2016: 779-788.

[23] Redmon J, Farhadi A. Yolov3: An incremental improvement[J]. arXiv preprint arXiv:1804.02767, 2018.

[24] Lin T Y, Goyal P, Girshick R, et al. Focal loss for dense object detection[C]//Proceedings of the IEEE international conference on computer vision. 2017. 2980-2988.

[25] Lin T Y, Dollár P, Girshick R, et al. Feature pyramid networks for object detection[C]//Proceedings of the IEEE conference on computer vision and pattern recognition. 2017: 2117-2125.

[26] Ge Z, Liu S, Wang F, et al. Yolox: Exceeding yolo series in 2021[J]. arXiv preprint arXiv:2107.08430, 2021.

[27] Hosang J, Omran M, Benenson R, et al. Taking a deeper look at pedestrians[C]//Proceedings of the IEEE conference on computer vision and pattern recognition. 2015: 4073-4082.

[28] Li J, Liang X, Shen S M, et al. Scale-aware fast R-CNN for pedestrian detection[J]. IEEE transactions on Multimedia, 2017, 20(4): 985-996.

[29] Pang Y, Xie J, Khan M H, et al. Mask-guided attention network for occluded pedestrian detection[C]//Proceedings of the IEEE/CVF international conference on computer vision. 2019: 4967-4975.

[30] Xu M, Bai Y, Qu S S, et al. Semantic Part RCNN for Real-World Pedestrian Detection[C]//CVPR Workshops. 2019.

[31] Huang X, Ge Z, Jie Z, et al. Nms by representative region: Towards crowded pedestrian detection by proposal pairing[C]//Proceedings of the IEEE/CVF Conference on Computer Vision and Pattern Recognition. 2020: 10750-10759.

[32] Hasan I, Liao S, Li J, et al. Generalizable pedestrian detection: the elephant in the room[C]//Proceedings of the IEEE/CVF Conference on Computer Vision and Pattern Recognition. 2021: 11328-11337.

[33] Li Q, Su Y, Gao Y, et al. OAF-Net: An Occlusion-Aware Anchor-Free Network for Pedestrian Detection in a Crowd[J]. IEEE Transactions on Intelligent Transportation Systems, 2022, 23(11): 21291-21300.

[34] Ren S, He K, Girshick R, et al. Faster r-cnn: Towards real-time object detection with region proposal networks[J]. Advances in neural information processing systems, 2015, 28.

[35] Zhang L, Lin L, Liang X, et al. Is faster R-CNN doing well

for pedestrian detection?[C]//Computer Vision-ECCV 2016: 14th European Conference, Amsterdam, The Netherlands, October 11-14, 2016, Proceedings, Part II 14. Springer International Publishing, 2016: 443-457.

[36] Brazil G, Yin X, Liu X. Illuminating pedestrians via simultaneous detection & segmentation[C]//Proceedings of the IEEE international conference on computer vision. 2017: 4950-4959.

[37] Hsu W Y, Lin W Y. Ratio-and-scale-aware YOLO for pedestrian detection[J]. IEEE transactions on image processing, 2020, 30: 934-947.

[38] Liu W, Liao S, Hu W, et al. Learning efficient single-stage pedestrian detectors by asymptotic localization fitting[C]//Proceedings of the European Conference on Computer Vision (ECCV). 2018: 618-634.

[39] Lin C, Lu J, Wang G, et al. Graininess-aware deep feature learning for pedestrian detection[C]//Proceedings of the European conference on computer vision (ECCV). 2018: 732-747.

[40] Tsotsos J K. A computational perspective on visual attention [M]. MIT Press, 2021.

[41] Hu J, Shen L, Sun G. Squeeze-and-excitation networks[C]//Proceedings of the IEEE conference on computer vision and pattern recognition. 2018: 7132- 7141.

[42] Woo S, Park J, Lee J Y, et al. Cbam: Convolutional block attention module[C]//Proceedings of the European conference on computer vision (ECCV). 2018: 3-19.

[43] Bello I, Zoph B, Vaswani A, et al. Attention augmented convolutional networks[C]//Proceedings of the IEEE/CVF international conference on computer vision. 2019: 3286-3295.

[44] Misra D, Nalamada T, Arasanipalai A U, et al. Rotate to attend: Convolutional triplet attention module[C]//Proceedings of the IEEE/CVF Winter Conference on Applications of Computer Vision. 2021: 3139-3148.

[45] Wang X, Girshick R, Gupta A, et al. Non-local neural networks[C]//Proceedings of the IEEE conference on computer vision and pattern recognition. 2018. 7794-7803.

[46] Liu J J, Hou Q, Cheng M M, et al. Improving convolutional networks with self-calibrated convolutions[C]//Proceedings of the IEEE/CVF conference on computer vision and pattern recognition. 2020: 10096-10105.

[47] Hadsell R, Chopra S, LeCun Y. Dimensionality reduction by learning an invariant mapping[C]//2006 IEEE Computer Society Conference on Computer Vision and Pattern Recognition (CVPR'06). IEEE, 2006, 2: 1735-1742.

[48] Chopra S, Hadsell R, LeCun Y. Learning a similarity metric discriminatively, with application to face verification[C]//2005 IEEE Computer Society Conference on Computer Vision and Pattern Recognition (CVPR'05). IEEE, 2005, 1: 539-546.

[49] He K, Fan H, Wu Y, et al. Momentum contrast for unsupervised visual representation learning[C]//Proceedings of the IEEE/CVF conference on computer vision and pattern recognition. 2020: 9729-9738.

[50] Chen T, Kornblith S, Norouzi M, et al. A simple framework for contrastive learning of visual representations[C]//International conference on machine learning. PMLR, 2020: 1597-1607.

[51] Li C, Li L, Jiang H, et al. YOLOv6: A single-stage object detection framework for industrial applications[J]. arXiv preprint arXiv:2209.02976, 2022.

[52] He K, Zhang X, Ren S, et al. Deep residual learning for image recognition[C]//Proceedings of the IEEE conference on computer vision and pattern recognition. 2016: 770-778.

[53] Simonyan K, Zisserman A. Very deep convolutional networks for large-scale image recognition[J]. arXiv preprint arXiv:1409.1556, 2014.

[54] Jia X, Zhu C, Li M, et al. LLVIP: A visible-infrared paired dataset for low-light vision[C]//Proceedings of the IEEE/CVF International Conference on Computer Vision. 2021: 3496-3504.

[55] Dollar P, Wojek C, Schiele B, et al. Pedestrian detection: an evaluation of the state of the art[J]. IEEE transactions on pattern analysis and machine intelligence, 2011, 34(4): 743-761.

[56] Bochkovskiy A, Wang C Y, Liao H Y M. Yolov4: Optimal speed and accuracy of object detection[J]. arXiv preprint arXiv:2004.10934, 2020.

[57] Zhang H, Cisse M, Dauphin Y N, et al. mixup: Beyond empirical risk minimization[J]. arXiv preprint arXiv:1710.09412, 2017.

[58] Dollár P, Appel R, Belongie S, et al. Fast feature pyramids for object detection[J]. IEEE transactions on pattern analysis and machine intelligence, 2014, 36(8): 1532-1545.

[59] Zhang S, Benenson R, Schiele B. Filtered channel features for pedestrian detection[C]//CVPR. 2015, 1(2): 4.

[60] Zhang S, Benenson R, Schiele B. Citypersons: a diverse dataset for pedestrian detection[C]//Proceedings of the IEEE conference on computer vision and pattern recognition. 2017: 3213-3221.

[61] Tian Z, Shen C, Chen H, et al. Fcos: Fully convolutional one-stage object detection[C]//Proceedings of the IEEE/CVF international conference on computer vision. 2019: 9627-9636.

[62] Chu X, Zheng A, Zhang X, et al. Detection in crowded scenes: one proposal, multiple predictions[C]//Proceedings of the IEEE/CVF Conference on Computer Vision and Pattern Recognition. 2020: 12214-12223.

[63] Wang Q, Wu B, Zhu P, et al. ECA-Net: Efficient channel attention for deep convolutional neural networks[C]//Proceedings of the IEEE/CVF conference on computer vision and pattern recognition. 2020: 11534-11542.

[64] Hou Q, Zhou D, Feng J. Coordinate attention for efficient mobile network design[C]//Proceedings of the IEEE/CVF conference on computer vision and pattern recognition. 2021: 13713-13722.